\documentclass[runningheads]{llncs}
\usepackage{graphicx}
\usepackage{cite}
\usepackage{graphicx}
\usepackage{amsmath,amssymb} % define this before the line numbering.
\usepackage{color}
\usepackage{multirow}
\usepackage{floatrow}
\begin{document}
\title{Targeted Kernel Networks: Faster Convolutions with Attentive Regularization} 
% Replace with your title

\titlerunning{Targeted Kernel Networks}
% Replace with a meaningful short version of your title
%
\author{Kashyap Chitta}%\orcidID{0000-0002-3891-3230}}
%
%Please write out author names in full in the paper, i.e. full given and family names. 
%If any authors have names that can be parsed into FirstName LastName in multiple ways, please include the correct parsing, in a comment to the volume editors:
%\index{Lastnames, Firstnames}
%(Do not uncomment it, because you may introduce extra index items if you do that, we will use scripts for introducing index entries...)
\authorrunning{K. Chitta}
% Replace with shorter version of the author list. If there are more authors than fits a line, please use A. Author et al.
%

\institute{The Robotics Institute, Carnegie Mellon University\\
\email{kchitta@andrew.cmu.edu}\\}
\maketitle              % typeset the header of the contribution
\begin{abstract}
We propose Attentive Regularization (AR), a method to constrain the activation maps of kernels in Convolutional Neural Networks (CNNs) to specific regions of interest (ROIs). Each kernel learns a location of specialization along with its weights through standard backpropagation. A differentiable attention mechanism requiring no additional supervision is used to optimize the ROIs. Traditional CNNs of different types and structures can be modified with this idea into equivalent Targeted Kernel Networks (TKNs), while keeping the network size nearly identical. By restricting kernel ROIs, we reduce the number of sliding convolutional operations performed throughout the network in its forward pass, speeding up both training and inference. We evaluate our proposed architecture on both synthetic and natural tasks across multiple domains. TKNs obtain significant improvements over baselines, requiring less computation (around an order of magnitude) while achieving superior performance.

\end{abstract}
\section{Introduction}
\label{sec_intro}
Convolutional Neural Networks (CNNs) have been largely responsible for the significant progress achieved on visual recognition tasks in recent years \cite{AlexNet,ResNet,DenseNet}. By sharing weights to be used by convolutional kernels across the entire spatial area of their input activations, CNNs use translated replicas of learned feature detectors, allowing them to translate knowledge about good weight values acquired at one position in an image to all other positions. This leads to translational equivariance-- a translated input to a convolutional layer will end up producing an identically translated activation after passing through it.

Though it works well in nearly all situations, it is possible for this 'knowledge translation' to be a double-edged sword. By sharing weights across the whole input, we bias the network to prioritize learning representations that would be useful over the entire image area. Due to this, it may have to compromise on learning some weights that are critical to the network's final objective, simply because these weights were useful only in a small area of the whole image. The possibility of this happening increases if the inputs possess a uniform spatial layout.

Assuming that the network inputs are captured or preprocessed in a way that provides some spatial structure, certain objects are more likely to be in particular locations than others. For example, if the inputs are all upright faces cropped with a face detector, it is far more likely to find an eye in one of the top quadrants of the input than in the bottom ones. In images of outdoor scenes, it is more likely to see blue skies at the top than the bottom. More often than not, there is some such spatial structure associated with the inputs to any visual recognition task. This means that based on what a kernel is supposed to look for, independently learning weights at different spatial locations can potentially generate better representations. 

A locally connected layer takes this idea to the extreme-- its forward pass involves convolutions with no weight sharing at all, with a different kernel for every spatial location in the input. By perfectly aligning facial images and then learning representations using locally connected layers, human-level accuracy was first achieved in face recognition \cite{DeepFace}. Unfortunately, the feasibility of this approach is limited due to the heavy dependence on perfect alignment of inputs and drastic increase in parameter count, leading to a requirement of far more training data (since there is no longer any 'knowledge translation').

Only sharing weights over selected Regions of Interest (ROIs) is another possibility that has been explored, implemented by training separate CNNs on different ROIs and merging their representations at a point deeper in the network \cite{DeepID1, DRML, EAC, ROINets}. The kernels are now specialized to their input ROIs, and the parameter count increase is controlled by architectural choices. Finding the right ROIs to use, however, is a tedious step usually requiring domain-specific knowledge to be done effectively. Any manual selection, even by the best domain experts, would almost certainly not lead to the most optimal choice of ROIs for the given task and network topology.

\begin{figure}[t!]
	\begin{center}
		\includegraphics[width=\textwidth]{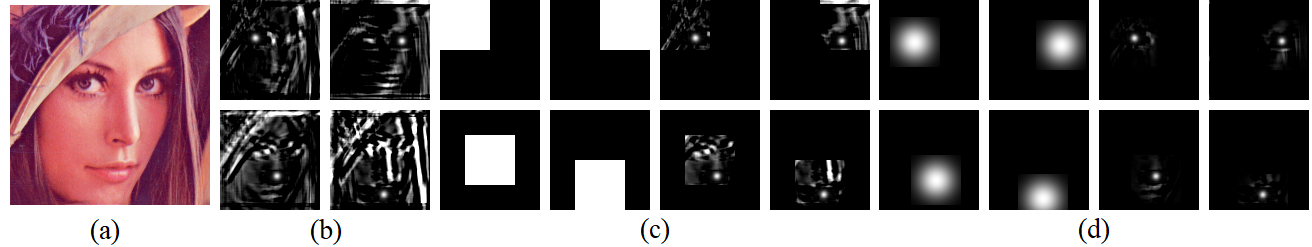}
	\end{center}
	\caption{(a) Input (b) Activations after a standard convolution with four kernels (actually correlation filters). These kernels are optimized to be activated by the left eye, right eye, nose and mouth respectively. However, they give large, unpredictable responses across the image. (c) Manual ROI selection and activations after convolutions on these selected ROIs. (d) The proposed approach, attention functions learned from data, and activations after AR. We observe that through spatial specialization, even crude features can become powerful, as they become independent of other spatial locations.}
	\label{f:pull}
\end{figure}

An alternative approach would be to learn the most optimal ROI for each kernel directly from the data, by end-to-end training. Trying to do this as a tuple of the ROI center and spatial size results in models that are not differentiable and require complex learning procedures \cite{DRAM}. We propose Attentive Regularization (AR), a method to achieve this using a differentiable attention mechanism \cite{DRAW}, allowing our models to be trained end-to-end with simple backpropagation. The key idea behind AR is to associate each rectangular ROI with the parameters of a smooth differentiable attention function. The attention function helps generate gradients of the loss with respect to the location and size of the ROI. Figure \ref{f:pull} illustrates the effect of AR, comparing it with a standard convolution and a fixed ROI based approach. For the purpose of illustration, we use a 'layer' operating on an RGB image with only four kernels, each looking for a semantically meaningful part.

An attractive consequence of having ROIs for each kernel is computational efficiency-- computing convolutions over small ROIs for every kernel in a layer greatly reduces redundant operations in the network, speeding up both training and evaluation.

\textbf{Our contribution} is three-fold: First, we propose and describe AR and its incorporation into existing CNN architectures, resulting in Targeted Kernel Networks (TKNs). Second, we evaluate TKNs on digit recognition benchmarks with coarse alignment in the form of digit centering, as well as synthetic settings with more alignment, significantly outperforming CNN baselines. Finally, we demonstrate their application for network acceleration on more complicated structured data, like faces and road traffic signs.

\section{Related Work}
\label{sec_lit}

\noindent
\textbf{Regularizing CNNs.} Deep CNNs have a vast potential to overfit data when they have to be trained from scratch. Conventional machine learning approaches to handle this like weight decay, data augmentation, and model ensembles alleviate the problem only to an extent. Dropout \cite{Dropout} was one of the most successful methods for regularizing layers with very large parameter counts in CNNs \cite{AlexNet,VGGNet}. 

Most recent models have substituted this with some constraint on the activations \cite{DARC}, the most popular of which is batch normalization \cite{BatchNorm}. This uses other images in the mini-batch along with learned scaling parameters to constrain the activations using computed statistics. We force the network to find good weights without giving the kernels free access to all spatial locations in the image during training, with a similar approach of applying constraints through learned parameters.

\noindent
\textbf{Spatially specialized CNNs.} Several approaches look into architectures that operate on ROIs, specifically in object detection \cite{RCNN, FastRCNN, FasterRCNN}. However, these methods typically propose ROI based object candidates for each input image, and not for the network kernels. Additional bounding box supervision is also necessary to learn these proposals. Unlike these methods, ROIs at a kernel level have been used in facial action unit detection \cite{FACS}, but the regions are hand-crafted \cite{DRML,ROINets,EAC}.

\noindent
\textbf{Attention.} One of the most promising trends in research is the emergence of attention based models. Early work in this area \cite{AncientAttention,RBMAttention1,RBMAttention2} was inspired by the process of sequential recognition used by the biological vision system in humans. Recent adaptations have leveraged the representational power of deep neural networks with visual attention for a variety of tasks, some of which were image classification \cite{SchmAttClass, Class1, Class2, Zoom}, image generation \cite{DRAW}, image captioning \cite{Captioning, Captioning2, Captioning3, Captioning4}, visual question answering \cite{VQA, VQA2, VQA3, VQA4}, action recognition \cite{AttPool} and one-shot learning \cite{ARC}. More closely related approaches to AR are attempts at multi-layer \cite{MulLay} and multi-channel \cite{Captioning3} attention. Our main advantage over existing soft attention methods is that we systematically remove computational processing throughout the network while maintaining the fully differentiable property. Other approaches require hard attention with reinforcement learning for network acceleration.

\noindent
\textbf{Efficient CNNs.} Cheng et al. \cite{Compression} summarize model compression and acceleration approaches into four categories-- parameter pruning and sharing \cite{Prune1, Prune2, Prune3, Prune4, Prune5}, low-rank factorization \cite{LRF1, LRF2, LRF3}, transferred or compact convolutional filters \cite{Trans1, Trans2, Trans3}, and knowledge distillation \cite{KD1,KD2,KD3,KD4}. One of the primary goals of early attention models was increasing efficiency \cite{DRAM}. This has resurfaced recently in the form of various architectures for spatially restricting computation.

\noindent
\textbf{Spatial Computation Restriction in CNNs.} Dynamic Capacity Networks \cite{DCN} define attention maps to apply sub-networks to only specific input patches for fine representations, which they later combine with the representations of a coarse network. Similarly, SBNet \cite{SBNet} uses a low resolution sub-network to obtain a computation mask for the main deep network. A more recent idea uses a learnable application of channel-wise sparsity to completely eliminate certain kernels dynamically \cite{MoreLess}. All these techniques restrict computation to the uncertain regions of the current image, whereas in our work, we restrict computation to certain (learnable) regions for all images. The two ideas are orthogonal and computational gains could be observed by combining them.

PerforatedCNNs \cite{PerfCNN} study strategies for skipping calculation of convolutions tied to certain spatial locations in a convolutional layer. These strategies are loosely based on using grid-like lattices, where computations at the intermediate points are approximated with interpolation. Our work removes computation in a similar fashion, but no interpolation is required since we do not have any intermediate values to recover.

\section{Attentive Regularization}
\label{sec_main}

In its simplest form, AR can be considered an additional layer operating on the activation of a convolutional layer using an attention window. We begin by explaining the one-dimensional implementation in this form before moving on to the generalized version and higher dimensional inputs.

\subsection{AR in One Dimension}
Consider the activation tensor $\textbf{A} \in \mathbb{R}^{D \times L}$ resulting from a one dimensional convolution of a sequence of length $L$ with $D$ different kernels. Let $\textbf{a}^{C} \in \mathbb{R}^{L}$ denote the row of this tensor corresponding to the $C^{th}$ kernel in the layer. The objective of AR is to constrain each one of these activation vectors using a differentiable attention function $f_{att}$. The window for attention is constructed as this function drops off numerically from $1$ to $0$. By sampling $f_{att}(x)$ at $L$ equally spaced points, we obtain an equivalent attention vector representing our function, $\textbf{f}_{att} \in \mathbb{R}^{L}$. Element-wise multiplication can now be used to weigh the original activation vector using its corresponding attention vector:
\begin{equation}
	\textbf{a}_{att}^{C} = \textbf{a}^{C} \odot \textbf{f}_{att}^{C}
	\label{e:one}
\end{equation}
where $\textbf{a}_{att}^{C}$ is the attentively regularized activation along the channel $C$, and $\odot$ denotes the element-wise product.

The key to optimizing the area of specialization of the kernels is now a problem of learning the right parameters to define the function $f_{att}$.

\subsection{Differentiable Functions for Attention}
The most obvious choice of $f_{att}$ to create a smooth attention window is the Gaussian function:
\begin{equation}
	f_{att}(x; \mu,\sigma) = e^{{{ - \left( {x - \mu } \right)^2 } \mathord{\left/ {\vphantom {{ - \left( {x - \mu } \right)^2 } {2\sigma ^2 }}} \right. \kern-\nulldelimiterspace} {2\sigma ^2 }}}
\end{equation}

This is completely parametrized by two variables, its mean $\mu$ and variance $\sigma^2$. Every time an update is applied to the convolutional layer weights during backpropagation, we can also update these two parameters in the AR layer. By varying $\mu$, the attention can translate to the optimal location in the sequence, and varying $\sigma^2$ allows the layer to learn the optimal scale, i.e., the amount of focus to pay at the chosen location.

\begin{figure}[t]
	\begin{center}
		\includegraphics[width=\textwidth]{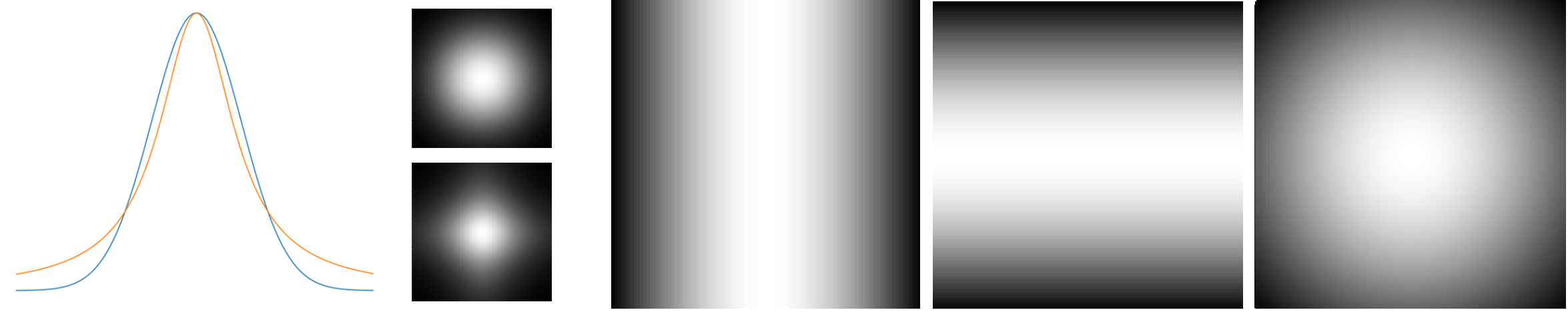}
	\end{center}
	\caption{Left: Gaussian (blue) and Cauchy (orange) attention functions and the equivalent bivariate functions (Gaussian on top). The Cauchy function has more weight in the tail of the distribution. Right: One slice of $\textbf{F}_x$, $\textbf{F}_y$ and $\textbf{F}_{att}$ associated with a single 2D kernel at initialization, using the Gaussian function as $f_{att}$. Due to linear seperability, AR can be trained extremely efficiently.}
	\label{f:cauchy}
\end{figure}

We also experimented with Cauchy functions, which have distinctively heavier tails than the corresponding Gaussians as shown in Figure \ref{f:cauchy}. We premised that this property would improve the gradient flow and help speed up the training of our layers, following \cite{ARC}. The Cauchy function with mean $\mu$ and scale parameter $\sigma$ is given by:

\begin{equation}
	f_{att}(x; \mu,\sigma) = \frac{1}{\left[1 + \left(\frac{x-\mu}{\sigma}\right)^2\right]}
\end{equation}

\subsection{AR in Two Dimensions}
The same logic used for one dimension can be generalized to images by considering two-dimensional attention maps associated with each kernel, instead of the attention vectors used for sequences. The input tensor $\textbf{A} \in \mathbb{R}^{C \times H \times W}$ has slices $\textbf{A}^{C} \in \mathbb{R}^{H \times W}$. We build the attention map by sampling $\textbf{F}_{att}^{C}$ from a bivariate $\mathcal{F}_{att}(x,y)$ along both dimensions. While using the Gaussian function, this now takes the form: 

\begin{equation}
	\mathcal{F}_{att}(x,y;\mu _x,\mu _y,\sigma _x,\sigma _y,\rho)=e^{-\alpha(x,y)}
\end{equation}
where
\begin{equation}
	\alpha(x,y) = (f(x))^2 - 2\rho f(x) f(y) + (f(y))^2 
\end{equation}
\begin{equation}
	f(x) = \frac{x-\mu _x%
	}{\sigma _x}
\end{equation}
\begin{equation}
	f(y) = \frac{y-\mu _y%
	}{\sigma _y}
\end{equation}

The attentively regularized activation $\textbf{A}_{att}^{C}$ is now obtained by the same procedure of element-wise multiplication as in Eq. (\ref{e:one}).

In our experiments, we found that the correlation parameter $\rho$ introduces an unnecessary degree of freedom to the attention map, as all scales and translations can be achieved by learning only $\mu _x, \mu _y, \sigma _x$ and $\sigma _y$. Setting $\rho = 0$ allows for more efficiency through a linearly separable implementation. Let the $i^{th}$ row of $\textbf{A}^{C}$ be denoted by $\textbf{a}^{(C,i,:)}$. We initially compute an intermediate activation $\textbf{A}^{C}_{int}$ by performing the following operation on all $i$ rows of $\textbf{A}^{C}$:

\begin{equation}
	\textbf{a}_{int}^{(C,i,:)} = \textbf{a}^{(C,i,:)} \odot \textbf{f}_{x}^{C}
\end{equation}
and then follow up with an operation on each of the $j$ columns of $\textbf{A}^{C}_{int}$ to get our final activation $\textbf{A}_{att}^{C}$:

\begin{equation}
	\textbf{a}_{att}^{(C,:,j)} = \textbf{a}_{int}^{(C,:,j)} \odot \textbf{f}_{y}^{C}.
\end{equation}
Here $\textbf{f}_{x} \in \mathbb{R}^{H}$ and $\textbf{f}_{y} \in \mathbb{R}^{W}$ are simply two separate one-dimensional attention vectors sampled from:

\begin{equation}
	f_{x}(x; \mu_x,\sigma_x) = e^{{{ - \left( {x - \mu_x } \right)^2 } \mathord{\left/ {\vphantom {{ - \left( {x - \mu_x } \right)^2 } {2\sigma_x ^2 }}} \right. \kern-\nulldelimiterspace} {2\sigma_x ^2 }}}
\end{equation}

\begin{equation}
	f_{y}(y; \mu_y,\sigma_y) = e^{{{ - \left( {x - \mu_y } \right)^2 } \mathord{\left/ {\vphantom {{ - \left( {x - \mu_y } \right)^2 } {2\sigma_y ^2 }}} \right. \kern-\nulldelimiterspace} {2\sigma_y ^2 }}}
\end{equation}
when using the Gaussian function.

\subsection{Tensor-Based Implementation}
While working with batch-sized tensors, it is more efficient to pre-compute the entire tensor $\textbf{F}_{att} \in \mathbb{R}^{C \times H \times W}$ directly from the parameter vector of means $\textbf{m} \in \mathbb{R}^{C}$ and the vector of scale parameters $\textbf{s} \in \mathbb{R}^{C}$ after using tile operations to broadcast them to the required dimensions. The combined tensor of all $C$ attention vectors $\textbf{f}_{att} \in \mathbb{R}^{C \times H}$ (or $\mathbb{R}^{C \times W}$) can be computed as:

\begin{equation}
	\textbf{f}_{att}(\textbf{x};\textbf{m},\textbf{s}) = e^{{{ - \left( {\textbf{x} - \textbf{m}} \right)^2 } \mathord{\left/ {\vphantom {{ - \left( {\textbf{x}-\textbf{m}} \right)^2 } {2\textbf{s}^2}}} \right. \kern-\nulldelimiterspace} {2\textbf{s}^2 }}}
	\label{e:f}
\end{equation}

Where \textbf{x} is a range vector ($0$ to $H$ or $0$ to $W$) scaled to lie in $[0,1]$. \textbf{m} is initialized to a vector with each entry $0.5$ so the window is initially centered. \textbf{s} is initialized to a vector of ones, such that the window tapers off from a value of $1$ at the center to $f(\sigma=1)$ at the image boundaries. For the two-dimensional case, $\textbf{f}_{x} \in \mathbb{R}^{C \times H}$ and $\textbf{f}_{y} \in \mathbb{R}^{C \times W}$ are computed as in Eq. (\ref{e:f}), broadcasted into three dimensions ($\mathbb{R}^{C \times H \times W}$), and $\textbf{F}_{att}$ is computed as
\begin{equation}
	\textbf{F}_{att} = \textbf{F}_{x} \odot \textbf{F}_{y}
\end{equation}

This is illustrated in Figure \ref{f:cauchy}. Every forward pass, an AR layer computes the element-wise product of its input and this attention function. After the backward pass, the function shifts slightly based on the updates to the vectors $\textbf{m}$ and $\textbf{s}$. The forward pass layer function is defined as:
\begin{equation}
	\textbf{A}_{att} = \textbf{A} \odot \textbf{F}_{att}.
\end{equation}

In this work, we limit ourselves to AR in two dimensions. Its extension to higher dimensions is trivial, using linearly separable one-dimensional attention windows along each input dimension.

\subsection{Efficient Convolutions with Targeting}
$\textbf{F}_{att}$ multiplicatively scales \textbf{A} in the forward pass. Over training, as the values in \textbf{m} and \textbf{s} change, a portion of the activation far enough away from the mean on the attention window gets scaled down to very small values. This effect is magnified when AR is used repeatedly, leading to a large number of near-zero activations through the network.

We exploit the fact that these activations are all located far from the mean, by performing the convolution operation for each kernel in only a rectangular ROI around the mean. This is mathematically equivalent to using an approximation to $\textbf{F}_{att}$ for AR, with values below a certain threshold clipped down to zero. We determine this ROI, given by its top-left and bottom-right coordinates:
\begin{align}
	\nonumber \textbf{roi}^{C} = &[(\textbf{m}_x - \frac{\textbf{s}_x}{\sqrt{2}})\times W; (\textbf{m}_y - \frac{\textbf{s}_y}{\sqrt{2}})\times H;\\
	&(\textbf{m}_x + \frac{\textbf{s}_x}{\sqrt{2}})\times W; (\textbf{m}_y + \frac{\textbf{s}_y}{\sqrt{2}})\times H].
\end{align}

This sliced ROI is used by a \textit{target} layer that efficiently performs the composite operation of both convolution and AR.
\begin{equation}
	\textbf{A}_{tar}^C[\textbf{roi}^C] = \textbf{A}^C[\textbf{roi}^C] * \textbf{K}^C.
\end{equation}
\begin{equation}
	\textbf{A}_{att} = \textbf{A}_{tar} \odot \textbf{F}_{att}
\end{equation}
where $\textbf{K}^C$ is the $C^{th}$ kernel in the target layer, \textbf{A} is the input activation, $\textbf{A}_{tar}$ is the intermediate result after the sliced convolution and $\textbf{A}_{att}$ is the layer output. * denotes the single channel 2D convolution operation.

During training, the values of \textbf{m} and \textbf{s} are clipped such that the size of the ROI never collapses to a value smaller than the kernel width. In addition, the overall ROI values are clipped so as to not exceed the boundaries of the input activation. At initialization, the ROI for all kernels is the entire input activation. 

In all our experiments, convolutions are done after the required amount of padding at the input boundaries so as to maintain constant spatial dimensions. We do not use an additive bias term in any convolutional layer. Our models were implemented with TensorFlow \cite{TF} and Keras \cite{Keras}.

\section{Experiments}
We empirically demonstrate the effectiveness of TKNs on four tasks: digit recognition on the MNIST \cite{MNIST} and SVHN \cite{SVHN} datasets, traffic sign recognition on the German Traffic Sign Recognition Benchmark \cite{GTSRB}, and facial analysis on the UNBC-McMaster Pain Archive \cite{PainArchive}. We also generate the tlMNIST dataset as a sanity check for TKNs, which aids us in understanding the visualizations of the learned attention mechanism. 

\subsection{Datasets}
\noindent
\textbf{MNIST.} The MNIST dataset contains 28$\times$28 grayscale images of handwritten numerical digits (0-9). The dataset is divided into 60,000 images for training and 10,000 for testing. The number of images per digit is not uniformly distributed. We perform no data augmentation or preprocessing except division of pixel values by 255 to place them in the range $[0,1]$.

\begin{figure}[t]
	\begin{center}
		\includegraphics[width=\textwidth]{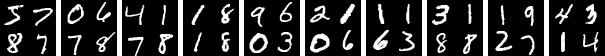}
	\end{center}
	\caption{tlMNIST data. The label assigned to each sample is that of the number in the top-left quadrant. The other three numbers serve as distractors to vanilla CNNs.}
	\label{f:tlMNIST}
\end{figure}

\noindent
\textbf{tlMNIST.} The tlMNIST dataset, short for top-left MNIST, is a set of 56$\times$56 grayscale images generated directly from MNIST. The 60,000 training images are created by placing each digit from the training partition of MNIST into the top-left 28$\times$28 quadrant of the images, and selecting 3 other digits from the same partition randomly to place in the other 3 image quadrants. The 10,000 image test set is similarly generated using only the test partition of MNIST. We use identical settings for both MNIST and tlMNIST experiments. The idea behind this task is to introduce a known synthetic 'alignment' to the data, so that it can be used as a sanity check for TKNs (kernels should focus on the top-left). Figure \ref{f:tlMNIST} shows some image samples from this dataset.

\noindent
\textbf{SVHN.} The SVHN dataset contains 32$\times$32 RGB digit images, cropped from pictures of house numbers. There are 73,257 images in the training set, 26,032 images in the test set, and 531,131 images for additional training. The digit of interest is centered in the cropped images, but nearby digits and other distractors are kept in the image. We train on only the 73,257 images in the training set, and report performances on the test set. Following \cite{WideResNet}, we do no preprocessing except pixel intensity scaling.

\noindent
\textbf{GTSRB.} GTSRB contains RGB images of road traffic signs taken in Germany, with bounding boxes provided for 43 different classes of signs. The main challenges of this dataset are low resolution and contrast. We follow the standard split for evaluation, involving 39,209 training images and a test set of 12,630 images. We preprocess each cropped bounding box by resizing it to $32\times32$, followed by pixel intensity scaling.

\noindent
\textbf{Pain.} The Pain Archive is a major publicly available test bed for research in facial analysis of induced pain expression. It consists of 200 video sequences of 25 subjects with 48,398 frames in total, each annotated with 66 facial landmarks and pain intensity levels (on a scale of 0-16). We split off around 30\% of the data (sequences of 7 of the subjects) for validation and use the remaining 70\% for training. This is a challenging task, which is also well suited to TKNs as we can preprocess the frames to create scale and viewpoint invariance. This is done by using the 66 landmark annotations to warp the faces to a frontal upright reference position before cropping and scaling to $48\times48$. We perform data augmentation by adding a small Gaussian noise to the landmarks before warping, and also randomly flipping the faces horizontally after warping.

\subsection{Training}
Our networks on the digit recognition tasks are trained using stochastic gradient descent (SGD). On MNIST and tlMNIST we train using batch size 128 for 20  epochs. The initial learning rate is set to 0.1, and is divided by 10 at the epochs 10 and 15. On SVHN, we train our models for 40 epochs with a batch size of 64. The learning rate is set to 0.1 initially, and is lowered by a factor of 10 after 20 epochs. Following \cite{DenseNet}, we use a weight decay of $10^{-4}$ and a Nesterov momentum \cite{Nesterov} of 0.9 without dampening.

On GTSRB and Pain, we use the Adam optimizer \cite{Adam} with a learning rate of 0.001, and train for a total of 100 epochs with a batch size of 64. For GTSRB, we use a higher weight decay of 0.05. We adopt the weight initialization introduced by He et al. \cite{HeInit}. We checkpoint the models after every epoch of training and report the error rates of the best single model. Test errors were only evaluated once for each task and model setting.

\subsection{Network Architectures}

\begin{table}[t!]
	\begin{center}
		\scriptsize
		\begin{tabular}{c|c|c|c|c|c}
			\hline
			\textbf{Layers} & \textbf{Output} & \textbf{CNN6} & \textbf{DN10} & \textbf{DN40} & \textbf{STN}\\
			\hline
			\multirow{3}{*}{Convolution (1)} & \multirow{3}{*}{$n\times n$} & \multirow{3}{*}{\textbf{5} $\times$ \textbf{5 conv}} & \multicolumn{2}{|c|}{3 $\times$ 3 conv} & LocNet \\
			\cline{4-6}
			& & & $ \text{\textbf{[3}}\times\text{\textbf{3 conv]}} \times$ 2 & $\left[ \begin{array}{cc} \text{1}\times\text{1 conv} \\ \text{\textbf{3}}\times\text{\textbf{3 conv}} \end{array}\right] \times$ 6 & \multirow{1}{*}{\textbf{7} $\times$ \textbf{7 conv}} \\
			\hline
			\multirow{2}{*}{Transition (1)} & \multirow{4}{*}{$\frac{n}{2}\times \frac{n}{2}$} & \multirow{2}{*}{2 $\times$ 2 max pool} & 1 $\times$ 1 conv& 1 $\times$ 1 conv, $\theta$ = 0.5 & \multirow{2}{*}{2 $\times$ 2 max pool}\\
			\cline{4-5}
			& & & \multicolumn{2}{|c|}{2 $\times$ 2 avg pool} & \\
			\cline{1-1} \cline{3-6}
			\multirow{1}{*}{Convolution (2)} & & \multirow{1}{*}{\textbf{5} $\times$ \textbf{5 conv}} & $ \text{\textbf{[3}}\times\text{\textbf{3 conv]}} \times$ 2 & $\left[ \begin{array}{cc} \text{1}\times\text{1 conv} \\ \text{\textbf{3}}\times\text{\textbf{3 conv}} \end{array}\right] \times$ 6 & \multirow{1}{*}{\textbf{5} $\times$ \textbf{5 conv}}\\
			\hline
			\multirow{2}{*}{Transition (2)} & \multirow{4}{*}{$\frac{n}{4}\times \frac{n}{4}$} & \multirow{2}{*}{2 $\times$ 2 max pool} & 1 $\times$ 1 conv & 1 $\times$ 1 conv, $\theta$ = 0.5 & \multirow{2}{*}{2 $\times$ 2 max pool}\\
			\cline{4-5}
			& & & \multicolumn{2}{|c|}{2 $\times$ 2 avg pool} & \\
			\cline{1-1} \cline{3-6}
			\multirow{1}{*}{Convolution (3)} & & \multirow{1}{*}{\textbf{5 $\times$ 5 conv}} & $ \text{\textbf{[3}}\times\text{\textbf{3 conv]}} \times$ 2 & $\left[ \begin{array}{cc} \text{1}\times\text{1 conv} \\ \text{\textbf{3}}\times\text{\textbf{3 conv}} \end{array}\right] \times$ 6 & \multirow{1}{*}{\textbf{3 $\times$ 3 conv}} \\
			\hline
			\multirow{2}{*}{Flatten} & \multirow{2}{*}{1$\times$1} & 328D FC& \multicolumn{2}{c|}{\multirow{2}{*}{$\frac{n}{4}\times \frac{n}{4}$ global avg pool}} & 2 $\times$ 2 max pool\\
			& & 192D FC & \multicolumn{2}{c|}{ } & 96D FC \\
			\hline
		\end{tabular}
	\end{center}
	\caption{CNN baselines. Convolutional layers replaced in TKNs are marked in \textbf{bold}. $\theta$ is the compression factor used to reduce the number of channels using a 1$\times$1 convolution at the transition blocks. The activation depths are reduced from $C$ to $(1-\theta)C$ at these layers. LocNet is a small localization network used to perform a learnable affine transform on the input. The final regression or classification layer is an FC layer with dimensions based on the task (10 for MNIST/SVHN, 43 for GTSRB, and 1 for Pain). The softmax cross entropy loss is used for classification, and a Euclidean loss is used for regression.}
	\label{t:arch}
\end{table}

To show that the benefits of AR are model-agnostic, we use four different CNN baselines across experiments. They are summarized in Table \ref{t:arch}. The first is a vanilla 6-layer CNN network with 3 convolutional layers (with 256, 256 and 128 kernels respectively) and 3 fully connected (FC) layers. The last layer is regularized with a dropout \cite{Dropout} of 0.5, and the ReLU non-linearity is used for all intermediate layers. %We denote this model as CNN6.

The second is a DenseNet \cite{DenseNet} with 3 densely connected blocks of 2 layers each. We use a growth rate of 12 and do not perform compression at the transition layers between blocks. We denote this model as DN10. Note that all convolutions in DenseNets are actually performed as the composite function, Batch Normalization \cite{BatchNorm} -- ReLU -- convolution.

We use a single baseline for our SVHN and Pain Archive experiments, a DenseNet-BC architecture with 3 blocks of 12 layers each. There are 21 connections in each block. We use a growth rate of 36, dropout with probability 0.2 after each convolution, and a compression factor of 0.5 at the 2 transition layers. We denote this model as DN40.

The final baseline is a Spatial Transformer Network (STN) \cite{STN} for GTSRB. This network learns how to warp the inputs with an affine transformation such that they are ideally aligned for the task. This meshes well with TKNs which are designed for aligned data. The main network we use is a 5-layer CNN with 3 convolutional layers (with 128, 128 and 256 kernels respectively) and 2 FC layers. We use batch normalization between all intermediate layers, and a dropout of 0.6 for the final FC layer. The localization network that computes warp parameters is a smaller version of the same network, with 3 convolutional layers of the same kernel size (with 16, 32 and 64 kernels respectively) and 3 FC layers (128, 64 and 6 units).

Given a CNN baseline, converting it to an equivalent TKN involves replacing convolutional layers with target layers. For the CNN6 and STN baselines, we simply replace all the convolutional layers in the main network, giving TKN6 and TSTN. For the DenseNet baselines, we replace the 3$\times$3 convolutional layers within the dense blocks, assuming that the bulk of the representation learning happens in these layers. We keep the initial, transitional and bottleneck 1$\times$1 convolutions as they are. We call these Targeted DenseNets (TDN10 and TDN40).

Further, there are three design choices within a target layer which we vary-- the choice of attention function ({Gaussian} or {Cauchy}); an $L_2$ weight penalty on scale parameters $\textbf{s}_x$ and $\textbf{s}_y$ to encourage more 'targeted' or 'focused' representations; and a multiplicative factor $\beta$ we build up the $L_2$ penalty by as we go deeper into the network, based on the intuition that deeper layers benefit less from weight sharing than shallow ones. This build up factor is applied by scaling the $L_2$ penalty by a factor of $\beta$ for all layers in the Convolution (2) block, and $\beta^2$ for all layers in the Convolution (3) block of the network.

\subsection{Results}

\begin{table}[t!]
	\begin{center}
		\begin{tabular}{l|c|c|c}
			\hline
			\textbf{Network} & \textbf{Params} & \textbf{FLOPs} & \textbf{Error}\\
			\hline
			Network in Network \cite{NIN} & - & - & 0.47 \\
			Deeply-Supervised Nets \cite{DSN} & - & - & 0.39 \\
			Competitive Multi-scale Convolution \cite{CMSC} & 4.48M & 632M & 0.33 \\
			CapsNet \cite{CapsNet} & 8.21M & 202M & {0.25} \\
			\hline
			CNN6 & 4.59M & 368M & 0.42\\
			\hline
			TKN6 (Gaussian, $L_2=10^{-4}, \beta=2$)& 4.59M & 52.9M & 0.48\\
			TKN6 (Cauchy, $L_2=10^{-4}, \beta=4$) & 4.59M & 28.6M & 0.43\\
			\hline
			DN10 & 44.7K & 11.3M & 0.48 \\
			\hline
			TDN10 (Gaussian, $L_2=10^{-4}, \beta=2$)& 45.0K & 6.93M & 0.42\\
			TDN10 (Cauchy, $L_2=10^{-4}, \beta=4$)& 45.0K & \textbf{6.26M} & \textbf{0.38}\\
			\hline
		\end{tabular}
	\end{center}
	\caption{Error rates (\%) on the MNIST dataset. Our best results in \textbf{bold}. AR improves both performance and efficiency.}
	\label{t:MNIST}
\end{table}

\begin{table}[t!]
	\begin{center}
		\begin{tabular}{l|c|c|c}
			\hline
			\textbf{Network} & \textbf{Params} & \textbf{FLOPs} & \textbf{Error}\\
			\hline
			CNN6 & 10.76M & 1470M & 0.83\\
			\hline
			TKN6 (Gaussian, $L_2=10^{-4}, \beta=2$)& 10.76M & 145M & 0.48\\
			TKN6 (Cauchy, $L_2=10^{-4}, \beta=2$) & 10.76M & 125M & 0.48\\
			TKN6 (Cauchy, $L_2=10^{-4}, \beta=4$) & 10.76M & 68.3M & 0.53\\
			\hline
			DN10 & 44.7K & 45.2M & 0.50 \\
			\hline
			TDN10 (Gaussian, $L_2=10^{-4}, \beta=2$)& 45.0K & 29.7M & 0.41\\
			TDN10 (Cauchy, $L_2=10^{-4}, \beta=2$)& 45.0K & \textbf{25.6M} & \textbf{0.38}\\
			\hline
		\end{tabular}
	\end{center}
	\caption{Error rates (\%) on the tlMNIST dataset. Our best results in \textbf{bold}. With AR, performance on tlMNIST becomes equivalent to the standard MNIST task.}
	\label{t:tlMNIST}
\end{table}

We compare our results on MNIST to other approaches that use single models with no data augmentation in Table \ref{t:MNIST}. Our best model does better than all previous CNN based methods on MNIST except a competitive multi-scale convolutional approach \cite{CMSC}. We are also outperformed by CapsNets \cite{CapsNet}, a new kind of neural network and not a drop in modification like AR. Both these network types have far more parameters and computational expenses than ours.

The TKNs corresponding to the CNN6 baseline (TKN6) match its performance, coupled with a huge boost in efficiency (\textbf{13$\times$ less floating point operations} in the forward pass). Introducing target layers benefits both the efficiency and performance when used with the DN10 models (\textbf{21\% reduced error rate}).

The results on tlMNIST are shown in Table \ref{t:tlMNIST}. Since the input data is highly 'aligned', we see significant improvement in results for both baselines. Another interesting observation is that the performance of the best networks on tlMNIST matches the MNIST results, showing that the effect of additional distractors has been completely negated by AR.

The results on SVHN are shown in Table \ref{t:SVHN}. Since models with the {Gaussian} attention function were far more difficult to tune in experiments on MNIST, we fix the {Cauchy} attention function for the remaining experiments. We obtain the best reported results (to our knowledge) on the reduced SVHN dataset where the extra training images are not used.

\begin{table}[t!]
	\begin{center}
		\begin{tabular}{l|c|c|c}
			\hline
			\textbf{Network} & \textbf{Params} & \textbf{FLOPs} & \textbf{Error}\\
			\hline
			CapsNet \cite{CapsNet} & 1.00M & 41.3M & {4.25} \\
			\hline
			DN40 & 0.83M & 357M & 3.17 \\
			\hline
			TDN40 (Cauchy, $L_2=10^{-4}, \beta=2$)& 0.83M & \textbf{205M} & \textbf{3.11}\\
			\hline
		\end{tabular}
	\end{center}
	\caption{Error rates (\%) on the SVHN dataset. Our best results in \textbf{bold}. We obtain state-of-the-art results on this reduced SVHN training set.}
	\label{t:SVHN}
\end{table}

The classification errors on the GTSRB test set are shown in Table \ref{t:GTSRB}. We also report the mean squared error (MSE) and mean average error (MAE) for regression on the validation partition of the Pain Archive in Table \ref{t:Pain}. On both tasks, we see distinctive benefits in terms of efficiency without loss in performance, showing the applicability of AR to network acceleration on practical tasks. Because we adopted hyper-parameter settings optimized for the CNN baselines in our study, we believe that further gains in accuracy of TKNs may be obtained by more detailed tuning of hyper-parameters and learning rate schedules.

\begin{table}[t!]
	\begin{center}
		\begin{tabular}{l|c|c|c}
			\hline
			\textbf{Network} & \textbf{Params} & \textbf{FLOPs} & \textbf{Error}\\
			\hline
			STN & 1.18M & 145M & 1.52\\
			\hline
			TSTN (Cauchy, $L_2=0.001, \beta=2$) & 1.18M & 55.7M & 1.53\\
			\hline
		\end{tabular}
	\end{center}
	\caption{Error rates (\%) on the GTSRB dataset. We achieve comparable performance with nearly a $3\times$ reduction in \#FLOPS.}
	\label{t:GTSRB}
\end{table}

\begin{table}[t!]
	\begin{center}
		\small
		\begin{tabular}{l | c| c | c |c }
			\hline
			\textbf{Network} & \textbf{Params} & \textbf{FLOPs} & \textbf{MSE} & \textbf{MAE} \\
			\hline
			DN40 & 0.83M & 802M & 1.67 & 0.51 \\
			\hline
			TDN40 (Cauchy, $L_2=10^{-4}, \beta=4$) & 0.83M & 391M & 1.67 & 0.50 \\
			\hline
		\end{tabular}
	\end{center}
	\caption{Regression errors (on a scale of 0-16) on the UNBC-McMaster Pain Archive. Here, we achive a $2\times$ reduction in \#FLOPS without loss in performance.}
	\label{t:Pain}
\end{table}

\begin{figure}[t]
	\begin{center}
		\includegraphics[width=\textwidth]{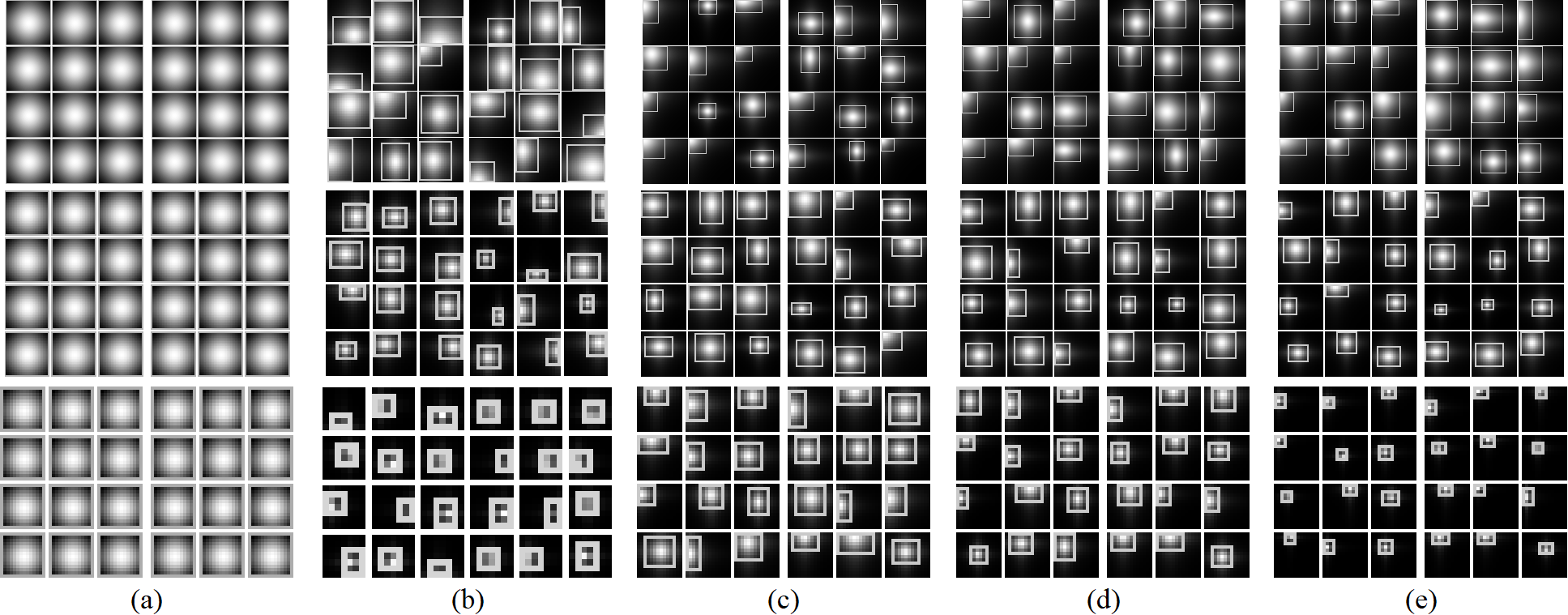}
	\end{center}
	\caption{Attention maps using the \textit{Cauchy} function. (a) Initialization. (b) After training on MNIST, $L_2=10^{-4}, \beta=4$. We notice large portions of the attention maps are vacant, particularly in the deeper layers. (c) After training on tlMNIST, $L_2=2\times10^{-4}, \beta=1$. (d) tlMNIST, $L_2=10^{-4}, \beta=2$. Though (c) and (d) have similar computational costs, (d) obtains slightly better performance. (e) tlMNIST, $L_2=10^{-4}, \beta=4$. Slightly better gains in efficiency can be obtained by scaling $\beta$ instead of $L_2$.}
	\label{f:vis}
\end{figure}

\section{Discussion}

Figure \ref{f:vis} shows the attention maps $\textbf{F}_{att}$ learned by the TDN10 models corresponding to each of the six target layers. Our experimental results combined with these visualizations give us some insight into the role of attention in CNN architectures. 

\noindent
\textbf{Implicit attention in CNNs.} A surprising observation is the near-identical error rate of the DN10 baseline on both MNIST (0.48\%) and tlMNIST (0.50\%). The network has no explicit way to pay more attention to any part of the input images, since it has no max pooling or FC layers. This means that for the tlMNIST task, the convolutional architecture itself learns to 'attend' to only the top-left portion of the image. This is possible because of the large convolutional receptive fields of the deeper layers. Each unit in the final convolutional layer has an effective receptive field larger than the entire input image. For tlMNIST, these units can learn locations associated with a given handwritten digit by simply looking for not just the digit, but a formation of the digit and a pattern associated with its location, such as the empty space to its bottom right and some portion of the three random digits around it. This is still an inconvenient task, which is why TKNs significantly improve the baseline (24\% reduction in error rate). The size of the receptive fields explains why the attention maps of the deeper layers in Figure \ref{f:vis} (c), (d) and (e) are not all on the extreme top-left portion of the image.

\noindent
\textbf{Fully Convolutional TKNs.} Each TKN kernel location is parametrized by $\textbf{m}_x$, $\textbf{m}_y$, $\textbf{s}_x$ and $\textbf{s}_y$, which are all {relative} values with respect to the absolute height and width of the image. Spatial structure in terms of layout is the crucial ingredient in the performance of TKNs, and if this remains similar, {they can be applied to images of varying sizes and aspect ratios} by using the same relative learned parameters scaled as per the new input resolution. To apply a TKN in a fully convolutional manner over a large image (for example, as a face detector), we first convert the relative parameters to absolute parameters by {choosing a scaling} for the attention layers in our fully convolutional TKN. This means a TKN learned at any resolution can be specialized to any other resolution by adjusting the chosen parameter scaling.

\noindent
\textbf{Network interpretability.} In traditional CNNs, we have seen how a deeper convolutional kernel may represent a mixture of patterns using its implicit attention and large receptive field. For example, when dealing with facial images, a kernel may learn to be activated by a certain combination of the eyes and mouth. Such complex knowledge representations greatly decrease the interpretability of the network \cite{Interp}. By introducing attention explicitly, kernels in TKNs can be encouraged to look at tiny areas in the inputs by increasing $L_2$. This makes them much more likely to be associated with single objects or parts, increasing the network interpretability. This is of great value when we need humans to trust a network's predictions.

\begin{figure}[t]
	\begin{center}
		\includegraphics[width=0.85\textwidth]{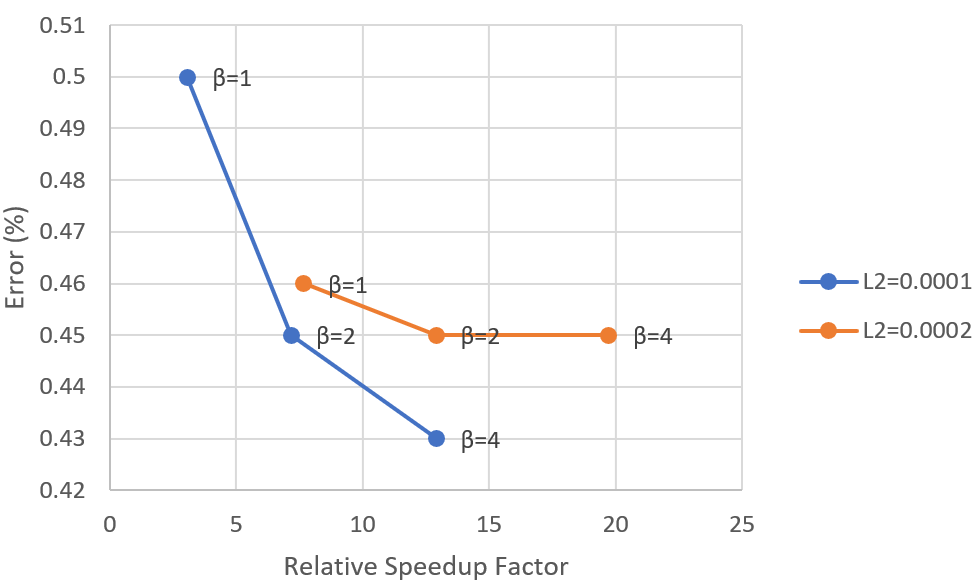}
	\end{center}
	\caption{Effect of $L_2$ and $\beta$ on performance and efficiency. Relative speedup factor is the ratio of \#FLOPs between the network and the CNN6 baseline. Better speed-performance tradeoffs are achieved through larger values of $\beta$.}
	\label{f:speed}
\end{figure}

\noindent
\textbf{Network acceleration.} Figure \ref{f:speed} shows the performance of TKN6 on MNIST as the $L_2$ penalty is varied. We see that a trade-off between speed and accuracy can be tuned by adjusting this penalty term while training. We also see that building up the $L_2$ penalty gradually over depth using $\beta$ improves performance in comparison to having a fixed penalty throughout the network. This validates our assumption that deeper, more abstract features require less weight sharing.

\section{Conclusion}
We proposed a new regularization method for CNNs called Attentive Regularization. It constrains the activation maps throughout the network to lie within specific ROIs associated with each kernel. This is done through a simple yet powerful modification of the convolutional layers, retaining end-to-end trainability with backpropagation. In our experiments, TKNs give a consistent improvement in efficiency over baselines in synthetic and natural settings, and competitive results to the state-of-the-art on benchmark datasets. Our experiments validate the idea that simplifying soft attention mechanisms to specific parametric distributions has potential for significant network acceleration.

In this study, we optimize for the attention parameters \textbf{m} and \textbf{s} for each kernel directly during training. In future work, we aim to study the effect of generating these parameters adaptively per image. Another extension to the proposed variant of TKNs would be to model the attention with a more complex function (like a mixture of Gaussians), or to use multiple kernels with different attention maps for the same output channel, making them deformable \cite{DPM}; to handle complex images where a single ROI per kernel may be insufficient.

\bibliographystyle{splncs04}
\bibliography{egbib}
\end{document}